\documentclass{article}

\PassOptionsToPackage{numbers, compress}{natbib}

    \usepackage[final]{paper_arxiv}

\usepackage[utf8]{inputenc} 
\usepackage[T1]{fontenc}    
\usepackage{hyperref}       
\usepackage{url}            
\usepackage{booktabs}       
\usepackage{amsfonts}       
\usepackage{nicefrac}       
\usepackage{microtype}      
\usepackage{xcolor} 
\usepackage{mathtools}
\usepackage{graphics}
\usepackage{graphicx}
\usepackage{wrapfig}
\usepackage[margins]{trackchanges}
\addeditor{jay}

\title{Geometric Priors for Scientific Generative Models in Inertial Confinement Fusion}

\author{%
  Ankita Shukla \\
  Arizona State University\\
  \texttt{ashukl20@asu.edu } \\
   \And
   Rushil Anirudh \\
   Lawrence Livermore National Laboratory\\
   \texttt{anirudh1@llnl.gov} \\
   \And 
   Eugene Kur \\
   Lawrence Livermore National Laboratory\\
   \texttt{kur1@llnl.gov} \\
   \And 
   Jayaraman J. Thiagarajan \\
   Lawrence Livermore National Laboratory\\
   \texttt{jjayaram@llnl.gov} \\
   \AND 
   Peer-Timo Bremer \\
   Lawrence Livermore National Laboratory\\
   \texttt{bremer5@llnl.gov} \\
   \And 
   Brian K. Spears \\
   Lawrence Livermore National Laboratory\\
   \texttt{spears9@llnl.gov} \\
   \And 
   Tammy Ma \\
   Lawrence Livermore National Laboratory\\
   \texttt{ma8@llnl.gov} \\
   \And 
   Pavan Turaga \\
   Arizona State University\\
   \texttt{pturaga@asu.edu} \\
}

\begin{document}

\maketitle

\begin{abstract}
In this paper, we develop a Wasserstein autoencoder (WAE) with a hyperspherical prior for multimodal data in the application of inertial confinement fusion. Unlike a typical hyperspherical generative model that requires computationally inefficient sampling from distributions like the von Mis Fisher, we sample from a normal distribution followed by a projection layer before the generator. Finally, to determine the validity of the generated samples, we exploit a known relationship between the modalities in the dataset as a scientific constraint, and study different properties of the proposed model.

\let\thefootnote\relax\footnote{This work was performed under the auspices of the U.S. Department of Energy by Lawrence Livermore National Laboratory under Contract DE-AC52-07NA27344}
\end{abstract}

\section{Introduction}
In this paper we explore the application of geometric priors inside generative models of scientific data -- particularly data obtained from extensive computational simulations of inertial confinement fusion~\cite{JAG_LLNL}. Such datasets present unique opportunities and challenges for generative models in that the underlying analytical generative process is typically known (for e.g., complex simulation codes), and there is a well understood demarcation between \emph{valid} and \emph{invalid} samples via known scientific constraints, laws etc. In this context, deep generative models present an exciting opportunity to explore the \emph{physics manifold} efficiently, to provide useful priors for solving challenging inverse problems, and to enable new insights into the underlying physics governing these datasets. Consequently, the past few years have witnessed several successful applications of generative models in a wide-range of physics problems ranging from inertial confinement fusion~\cite{anirudh2020improved} to cosmology~\cite{mustafa2019cosmogan} and high-energy particle physics~\cite{paganini2018calogan}. 

A common assumption in most of these cases is that the underlying geometry of the physics manifold is Euclidean, whereas this is often not the case in scientific or non-scientific datasets. This has led to the development of geometry-driven approaches that have improved the understanding, interpretability as well as the performance of deep models~\cite{defferrard2016convolutional,Cohen2018spherical,s-vae18,nickel2017poincare}. Many approaches exploit geometry of the data to inform design choices of neural networks -- molecules as graphs~\cite{defferrard2016convolutional}, or spherical images from drones or robots~\cite{Cohen2018spherical}. Furthermore, several studies have also focused on building better latent spaces in deep models via geometric constraints. These have been especially helpful in cases where the latent manifold of the data (but not the observed data itself) is known to come from a non-Euclidean manifold such as in the case of spherical \cite{s-vae18} or hyperbolic spaces \cite{nickel2017poincare,Wilson_TPAMI2014, Khrulkov_CVPR2020}. While the geometry of the dataset is explicitly known in the former, in the latter setting, the latent space is assumed to be a curved manifold with the expectation that the curvature of the latent space matches the true structure in the data. With generative models in particular, the Euclidean family of approaches tend to be efficient to train, and sample from, while their non-Euclidean counterparts are more accurate in capturing the underlying geometry though typically being more computationally involved. 

In this work, we investigate an intermediate approach that places a \emph{geometric prior} in the generator, without enforcing any kind of geometric constraint on the latent space. We find that this simple re-parameterization allows us to place geometric constraints inside the generative model while still using a prior distribution in the Euclidean space. Specifically, we study the hyper-sphere as the manifold of choice, due to several reasons -- it has been shown to have natural regularizing properties \cite{s-vae18}, it is relatively easy to incorporate in existing neural networks, and the fact that a normal distribution in the Euclidean space when projected onto a hyper-sphere becomes uniformly distributed on its surface. In this regard, we extend the Wasserstein autoencoder (WAE) \cite{tolstikhin2018wasserstein} with such a re-parameterization. In the standard WAE, the latent space is encouraged to match a prior, typically a normal distribution, during training. At inference time, samples from the prior are passed through the generator to obtain samples from the data distribution. Our approach places an explicit projection layer in the generator following the latent space. This allows us to place a hyper-spherical constraint while still allowing us to sample efficiently from a standard normal distribution.

\paragraph{Inertial Confinement Fusion} We use ICF as a test-bed for our analysis. ICF \cite{lindl1995development} uses powerful lasers to heat and compress a millimeter-scale target filled with thermonuclear deuterium-tritium (DT) fuel. The goal is to drive fusion reactions that self-heat the DT fuel leading to ignition and propagating burn.  A variety of diagnostic instruments are used to observe the fusion implosions that can be used to understand and drive the next set of experiments -- these range from x-ray and neutron cameras that create spectrally resolved (i.e. hypercolor) images, as well as a host of spectrometers and radiochemical diagnostics that produce key scalar features. As such, each sample in our dataset consists of a set of these multi-modal diagnostic quantities. As a notion of scientific validity, we use thermal equilibrium to relate predicted ion temperatures (as part of the scalars) and estimates of electron temperature formed by ratios of x-ray image brightnesses. As shown in the experiments, this manifests as a strong linear relationship between the predicted scalar, and the mean brightness of the x-ray images. Considering that the data from actual experiments tend to be highly sparse due to the cost of running multiple experiments, we restrict our analysis to a simulated ICF dataset. This forms a 1D semi-analytic simulator that outputs four images as well as a set of scalars that can be seen as samples from a high dimensional distribution.



\section{Experiments and Results}
\subsection{Dataset details}
We use the JAG Inertial Confinement Fusion Dataset \cite{JAG_LLNL} with 100K samples for our experiments. Each sample consists of 4 X-ray images of size $64 \times 64$, where each image corresponds to a measurement at a pre-defined energy. The 4 images are treated as channels, resulting in volume of $64 \times 64 \times 4$. Along with the images, there are 15 scalars that measure quantities such as yield, ion-temperature, pressure and other physical quantities. One of the scalars (ion temperature) is directly correlated to the image temperature (integral of the image) in that they are both independent measurements of the same quantity. We exploit this fact as a scientific prior to determine the validity of the generated samples.

\subsection{WAE with a hyper-spherical Prior}
We explore the use of a hyper-spherical embedding inside the generator while keeping the latent space Euclidean. This allows us to place a prior over the latent space using a standard normal distribution which is easy to sample from, as opposed to more complex distributions such as the von Mises-Fisher (vMF). In doing so, we exploit the fact that a normal distribution projected onto to a hyper-sphere forms a uniform distribution on its shell. We demonstrate a proof of concept by extending the Wasserstein autoencoder (WAE)~\cite{tolstikhin2018wasserstein} with an additional layer that projects the latent vector obtained from the encoder onto a unit hyper-sphere. 

We define a unit hyper-sphere as $\mathcal{S}^{d-1} = \{\textbf{x} \in \mathbb{R}^d: ||\mathbf{x}||_2=1\}$. Our proposed model consists of the following: an encoder network, $\mathcal{E}: (\mathbf{I},\mathbf{S}) \rightarrow \mathbf{z}$; where $\mathbf{I}$ denotes a multi-channel spectral image and $\mathbf{S}$ denotes a set of scalar diagnostics such as temperature, energy yield, pressure etc. To account for the multi-modal nature of the data, we use convolutional layers to process the image channels, and a fully connected layer to process the scalars. These are then stacked after a few layers before jointly predicting the bottleneck representation, $\mathbf{z}$.  Next, we define a projection map $\mathcal{P}: \mathbf{z} \rightarrow \mathcal{S}^{d-1}$, followed by a generator network $\mathcal{G}: S^{d-1} \rightarrow (\mathbf{I}, \mathbf{S})$. Here the projection map $\mathcal{P}$ is defined as $\mathbf{\tilde{z}}= \mathcal{P}(\mathbf{z})= \mathbf{z}/||\mathbf{z}||_2$. Thus, the encoder-decoder pair mapping is given by: $(\mathbf{I},\mathbf{S}) \xrightarrow[]{\mathcal{E}} \mathbf{z} \xrightarrow[]{\mathcal{P}} \mathbf{\tilde{z}} \xrightarrow[]{\mathcal{G}} (\mathbf{I},\mathbf{S})$.

We keep the discriminator from the standard Euclidean WAE unchanged, and use an adversarial objective to get the latent space $\mathbf{z}$ to match the prior, which in our case is the zero mean unit variance normal distribution, $\mathcal{N}(\mathbf{0},\mathbf{\mathrm{I}})$. During training, there are two key procedures that need to be taken into account--the generator is only fed samples from the hyper-sphere and the discriminator samples from the Euclidean Gaussian prior. This makes our approach simple to optimize as opposed to models with explicit hyper-spherical distribution. At test time, we simply sample from the prior and pass it along to the generator to create new image and scalar data.

We used the same network architecture as in \cite{anirudh2020improved}, with additional batch norm layers and 16 dimensional latent space. We trained the model for 600 epochs using Adam optimizer with $\beta_1= 0.5$ and $\beta_2 =0.999$ and learning rate $1e-3$. All the experiments run on a 22 GB, Quadro RTX 6000 GPU. 

\subsection{Evaluating the reconstructions with the scientific prior}
\begin{figure}[h]
    \centering
    \includegraphics[scale=0.15]{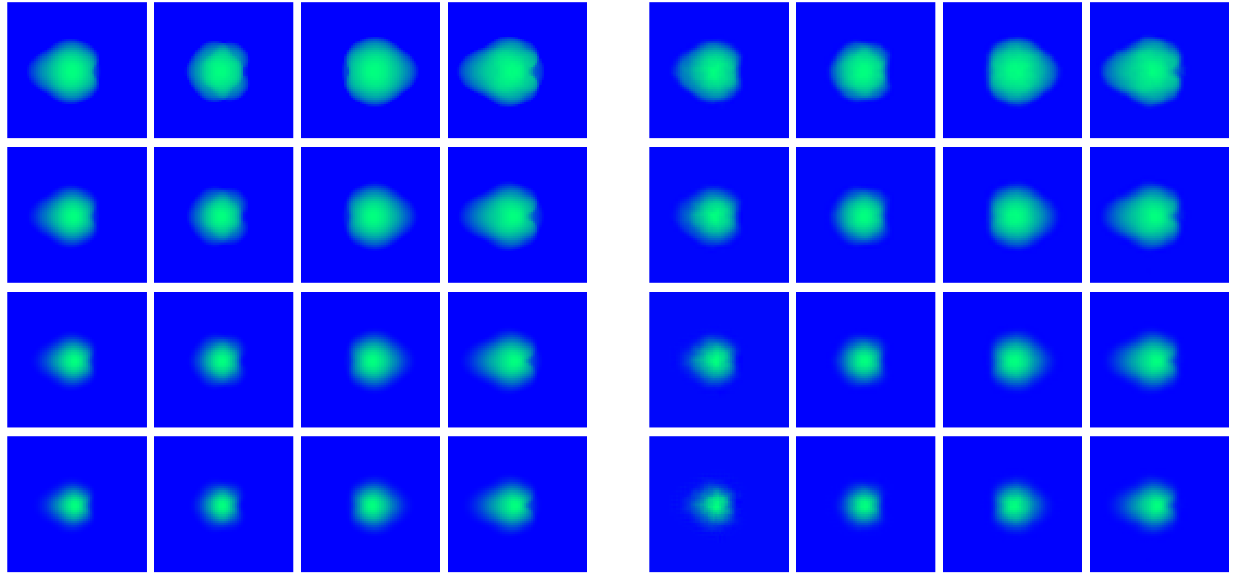}
    \includegraphics[scale=0.155]{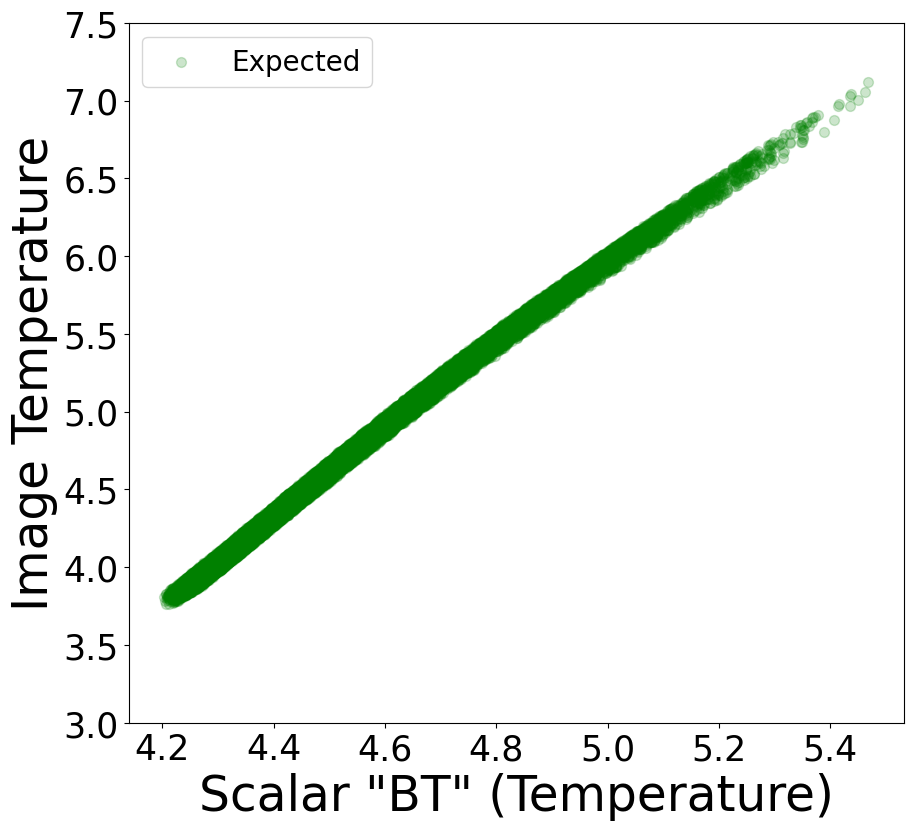}
    \includegraphics[scale=0.155]{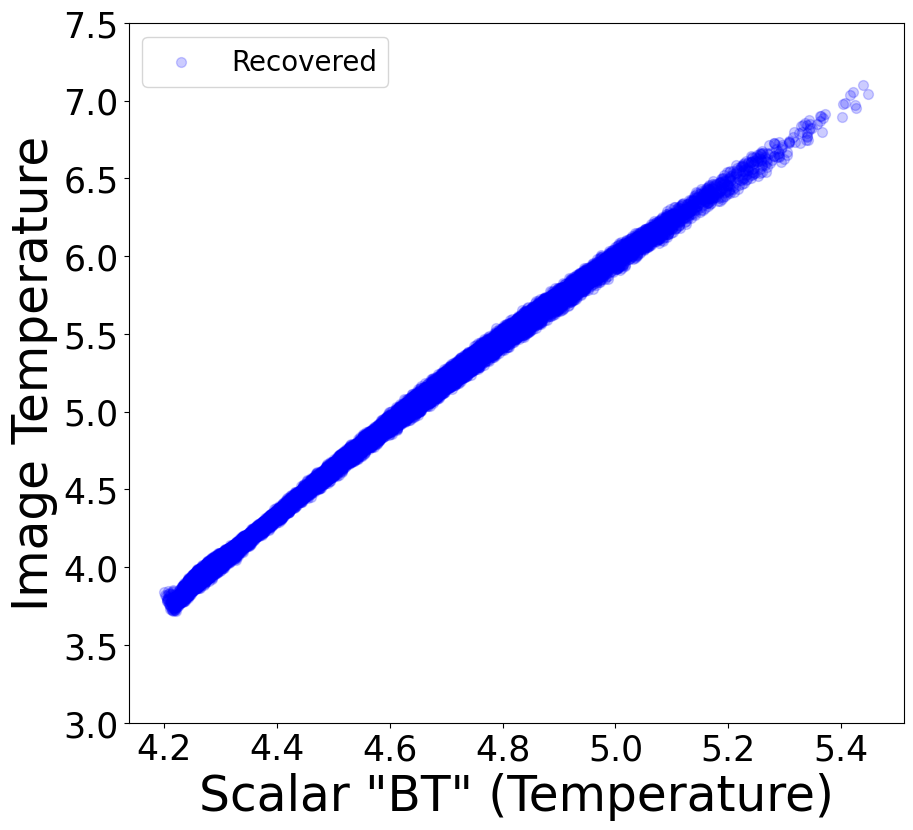}
    \caption{Example of original data (left) and the corresponding reconstructed data from the trained hyper-spherical parametrized WAE model along with their scientific priors (right).}
    \label{fig:recon_images}
\end{figure}

We evaluate the quantitative performance of proposed model using mean squared error (MSE) and  R-squared statistic ($R^2$) for images and 15 scalars respectively on the test set. Our approach achieves MSE of approximately 0.002 and $R^2$ score of approximately 0.99. We show a few sample images and their reconstructions from the four channels (each row) in the Figure \ref{fig:recon_images}. We validate the reconstructions of test samples further using the scientific prior as shown in the right panel of Figure \ref{fig:recon_images}, where each sample is represented as a point on the line. We see that the relationship between the images and scalars is preserved extremely well demonstrated by the strong linear relationship between the estimated temperature from the images and the temperature scalar.

\begin{figure}
    \centering
    \includegraphics[scale=0.21]{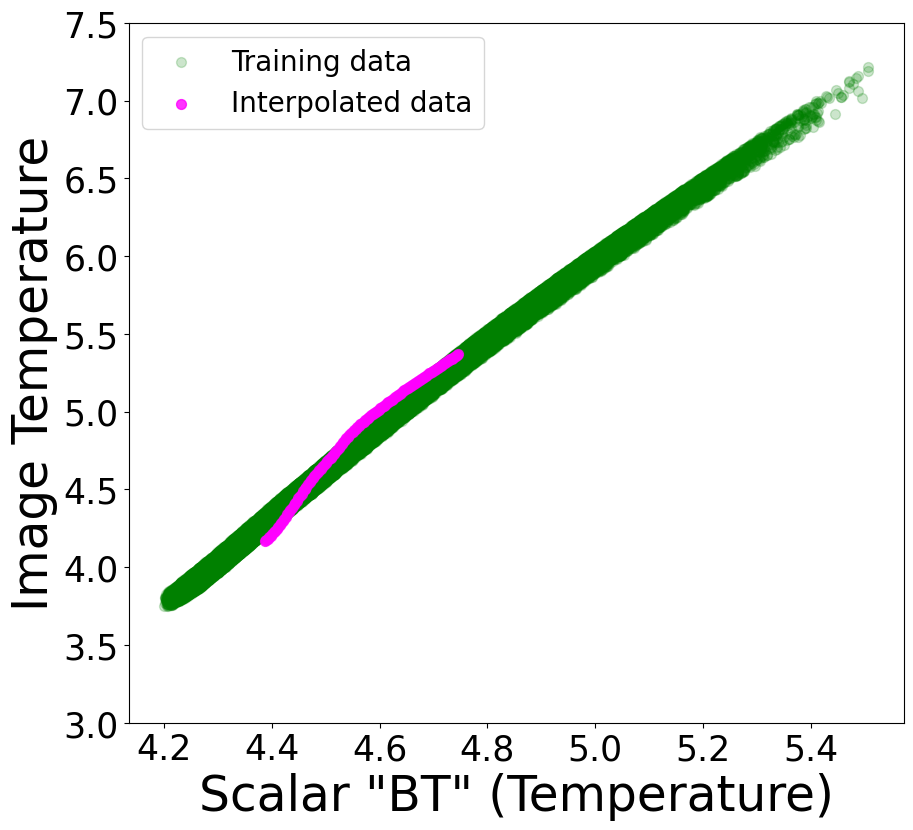}\includegraphics[scale=0.11]{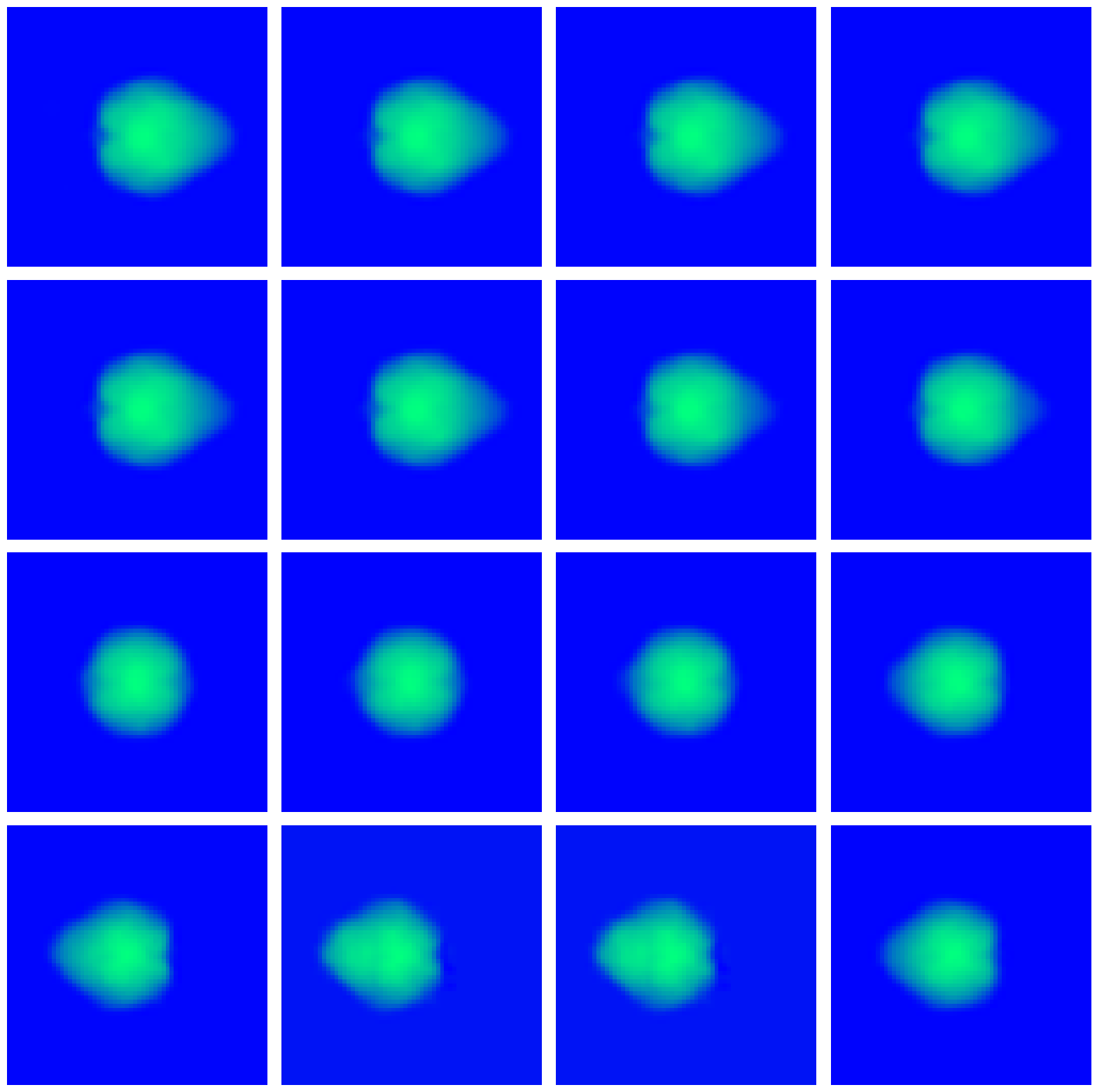}
    \caption{\textbf{Left:} Interpolating between two points in the latent space results in interpolation in 1D scientific prior manifold. \textbf{Right:} The corresponding images (only one channel is shown here) obtained from interpolation between the top left as the starting and bottom right as the end point. }
    \label{fig:interpolate}
\end{figure}
\paragraph{Interpolation in the latent space.}
We evaluate the continuity of the learned latent space, by performing interpolation between latent embeddings of training data. Since it is Euclidean, we perform linear interpolation between any two given training samples. We observe that the generated samples also satisfy the scientific constraint, indicating that the path taken by the geodesic is reflective of the true path on the physics manifold. We sample a few images from this path in the rightmost panel of Figure \ref{fig:interpolate}, where top left shows the starting point and the bottom right is the end point. As expected, we see that the learned latent space captures the semantic information in the dataset accurately -- for e.g., indicated by the direction of the tail in the image.

\begin{figure}
    \centering
    \includegraphics[scale=0.18]{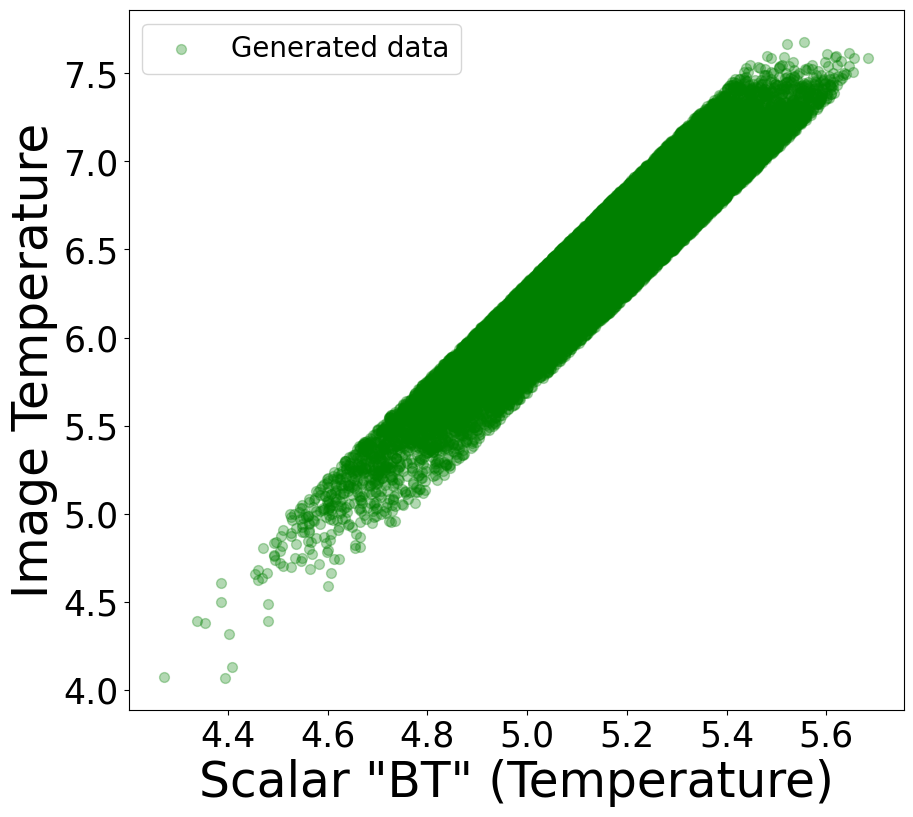} \hspace{0.2cm}
    \includegraphics[scale=0.18]{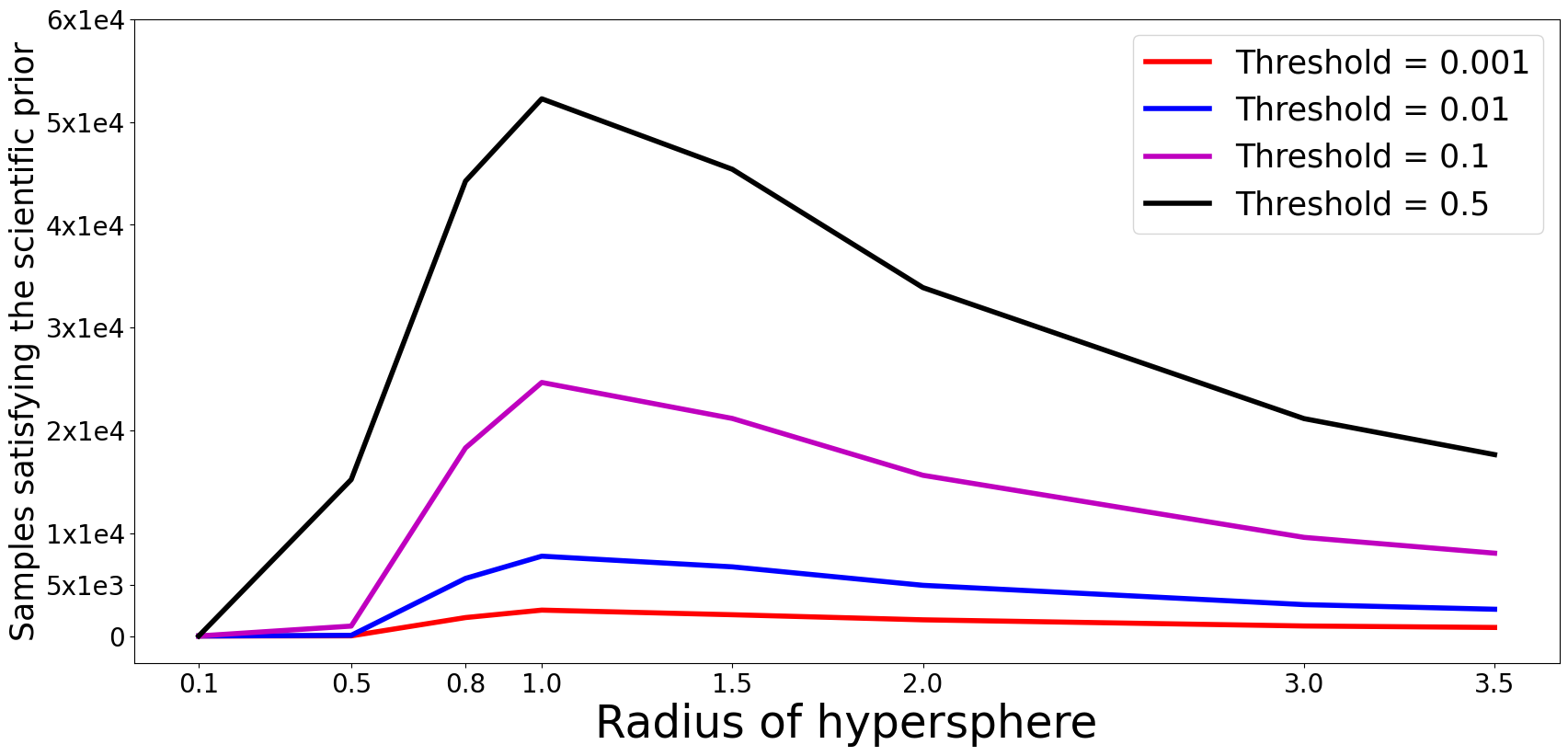}
    \caption{Evaluating the scientific prior on  generated data (left) and variation in number of valid generated samples based on deviation from the scientific prior for different threshold values and different radius of hyper-sphere (right).  }
    \label{fig:cons_satisfy}
\end{figure}

\paragraph{Evaluating scientific validity of generated samples.}
Once the WAE is trained, we sample from the prior and pass them through the generator to create new samples from the ICF dataset. To assess the scientific validity of these samples, we compute the scientific constraint on each one of these samples. Since the generator is not conditioned to follow the scientific prior, generated samples do not necessary result in valid simulation data. We observe that only a subset of samples satisfy the scientific constraint. Figure \ref{fig:cons_satisfy} shows evaluated scientific prior on generated samples.

\begin{figure}
    \centering
    \includegraphics[scale=0.25]{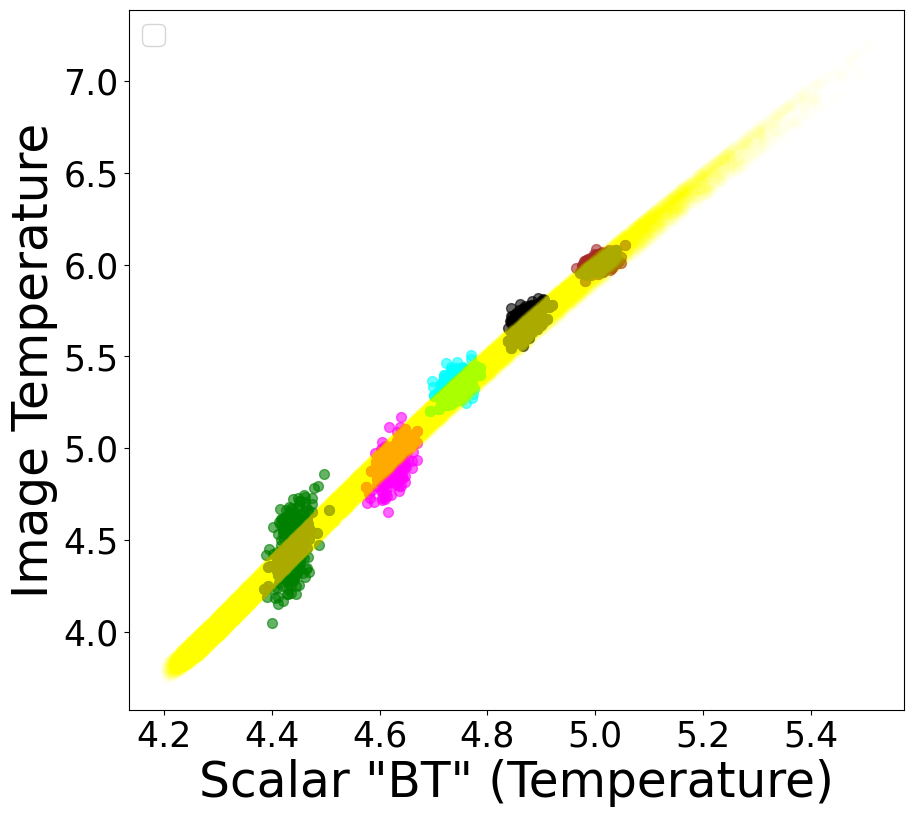}
    \caption{Sampling in the neighbourhood of training samples, generates samples that follow the scientific prior, the reduced variance indicates that the training samples are better sampled on the top end of the curve.}
    \label{fig:sampling}
\end{figure}

\paragraph{Effect of the hyper-sphere radius}
As all the samples from the prior undergo hyper-spherical parametrization, we obtain invariance w.r.t to the spread of the normal distribution, \emph{i.e}., sampling from $\mathcal{N}(0,k)$, results in the same distribution on the hypersphere. This sampling strategy indicates that the proposed model does not compromise on the diversity of generated samples, as opposed to a standard WAE, where there is typically a trade-off between diversity and sample quality depending on the variance of the normal distribution. 

Next, we study the effect of changing the radius of the hypersphere at test time on sample quality using the scientific constraint. Since, we parametrize the latent space on a unit hypersphere, smaller as well as larger radii generate fewer valid samples. We ablate on this in Figure \ref{fig:cons_satisfy} for four different threshold values (as a measure of valid vs invalid samples). Here the threshold is used to determine the deviation from the line fit on the training data. Finally, in Figure \ref{fig:sampling}, we sample around five different points using them as the mean in a normal distribution with unit variance in the latent space. By mapping the generated samples onto the scientific constraint, we see that they manifest with different spreads around the the training sample -- we hypothesize that these differences suggest irregularities in the training distribution.

\subsection{Discussion}
In this paper, we study hyper-spherical models for inertial confinement fusion. We modify the Wasserstien autoencoder (WAE) by including an additional normalizing step that maps the latent space onto a unit hyper-sphere. This presents a proof of concept for incorporating geometric priors (hyper-spherical, hyperbolic, or mixed curvature) into generative models for scientific datasets. Further, these models may also help better determine the geometry and curvature of the underlying physics manifold.

\small{\bibliographystyle{ieee}
\bibliography{ref}

\begin{thebibliography}{10}\itemsep=-1pt

\bibitem{anirudh2020improved}
R.~Anirudh, J.~J. Thiagarajan, P.-T. Bremer, and B.~K. Spears.
\newblock Improved surrogates in inertial confinement fusion with manifold and
  cycle consistencies.
\newblock {\em Proceedings of the National Academy of Sciences},
  117(18):9741--9746, 2020.

\bibitem{Cohen2018spherical}
T.~S. Cohen, M.~Geiger, J.~Köhler, and M.~Welling.
\newblock Spherical {CNN}s.
\newblock In {\em International Conference on Learning Representations}, 2018.

\bibitem{s-vae18}
T.~R. Davidson, L.~Falorsi, N.~De~Cao, T.~Kipf, and J.~M. Tomczak.
\newblock Hyperspherical variational auto-encoders.
\newblock {\em 34th Conference on Uncertainty in Artificial Intelligence
  (UAI-18)}, 2018.

\bibitem{defferrard2016convolutional}
M.~Defferrard, X.~Bresson, and P.~Vandergheynst.
\newblock Convolutional neural networks on graphs with fast localized spectral
  filtering.
\newblock {\em Advances in neural information processing systems},
  29:3844--3852, 2016.

\bibitem{JAG_LLNL}
J.~A. Gaffney, R.~Anirudh, P.-T. Bremer, J.~Hammer, D.~Hysom, S.~A. Jacobs,
  J.~L. Peterson, P.~Robinson, B.~K. Spears, P.~T. Springer, J.~J. Thiagarajan,
  B.~Van~Essen, and J.-S. Yeom.
\newblock The {JAG} inertial confinement fusion simulation dataset for
  multi-modal scientific deep learning.
\newblock In Lawrence Livermore National Laboratory (LLNL) Open Data
  Initiative. UC San Diego Library Digital Collections., 3 2020.

\bibitem{Khrulkov_CVPR2020}
V.~Khrulkov, L.~Mirvakhabova, E.~Ustinova, I.~Oseledets, and V.~Lempitsky.
\newblock Hyperbolic image embeddings.
\newblock In {\em Proceedings of the IEEE/CVF Conference on Computer Vision and
  Pattern Recognition}, pages 6418--6428, 2020.

\bibitem{lindl1995development}
J.~Lindl.
\newblock Development of the indirect-drive approach to inertial confinement
  fusion and the target physics basis for ignition and gain.
\newblock {\em Physics of plasmas}, 2(11):3933--4024, 1995.

\bibitem{mustafa2019cosmogan}
M.~Mustafa, D.~Bard, W.~Bhimji, Z.~Luki{\'c}, R.~Al-Rfou, and J.~M. Kratochvil.
\newblock Cosmogan: creating high-fidelity weak lensing convergence maps using
  generative adversarial networks.
\newblock {\em Computational Astrophysics and Cosmology}, 6(1):1, 2019.

\bibitem{nickel2017poincare}
M.~Nickel and D.~Kiela.
\newblock Poincar{\'e} embeddings for learning hierarchical representations.
\newblock {\em Advances in neural information processing systems},
  30:6338--6347, 2017.

\bibitem{paganini2018calogan}
M.~Paganini, L.~de~Oliveira, and B.~Nachman.
\newblock Calogan: Simulating 3d high energy particle showers in multilayer
  electromagnetic calorimeters with generative adversarial networks.
\newblock {\em Physical Review D}, 97(1):014021, 2018.

\bibitem{tolstikhin2018wasserstein}
I.~Tolstikhin, O.~Bousquet, S.~Gelly, and B.~Schoelkopf.
\newblock Wasserstein auto-encoders.
\newblock In {\em International Conference on Learning Representations}, 2018.

\bibitem{Wilson_TPAMI2014}
R.~C. Wilson, E.~R. Hancock, E.~Pekalska, and R.~P. Duin.
\newblock Spherical and hyperbolic embeddings of data.
\newblock {\em IEEE transactions on pattern analysis and machine intelligence},
  36(11):2255--2269, 2014.

\end{thebibliography}
}

\end{document}